\documentclass{article}

\usepackage[numbers]{natbib}
\usepackage[preprint]{neurips_2024}

\usepackage[dvipsnames]{xcolor}         
\definecolor{linkColor}{rgb}{0.18,0.39,0.62}
\usepackage[utf8]{inputenc} 
\usepackage[T1]{fontenc}    
\usepackage[colorlinks=true,linkcolor=linkColor,citecolor=linkColor,filecolor=linkColor,urlcolor=linkColor]{hyperref}       
\usepackage{url}            
\usepackage{booktabs}       
\usepackage{amsfonts}       
\usepackage{nicefrac}       
\usepackage{microtype}      

\usepackage{graphicx}
\usepackage{arydshln}

\usepackage{caption}
\usepackage{subcaption}
\usepackage{makecell}
\usepackage{csquotes}
\usepackage{epigraph}
\usepackage{amssymb}
\usepackage{mathtools}
\usepackage{amsthm}
\usepackage{algorithm}
\usepackage{algpseudocode}
\usepackage{algorithmicx}
\usepackage{amsmath}
\usepackage{multirow}
\usepackage{pifont}
\usepackage{mathtools}
\usepackage{setspace}
\usepackage{soul}
\usepackage{graphicx}
\usepackage{wrapfig}
\usepackage{makecell}
\usepackage[customcolors,norndcorners]{hf-tikz}

\usepackage{multirow}
\usepackage{amsmath}
\usepackage{capt-of}
\usepackage{tabularx}
\usepackage{epsfig}
\usepackage{amssymb}
\usepackage{amsfonts}
\usepackage{booktabs}
\usepackage{scalerel}
\usepackage[inline]{enumitem}
\usepackage{listings}
\usepackage{varwidth}
\usepackage[export]{adjustbox}
\usepackage{tikz}
\usetikzlibrary{tikzmark}

\usepackage{stmaryrd}
\usepackage{bbm}
\usepackage{wrapfig}
\usepackage{pifont}
\usepackage[noabbrev]{cleveref}

\definecolor{deepblue}{rgb}{0,0,0.5}
\definecolor{officeblue}{RGB}{0,102,204}
\definecolor{deepred}{rgb}{0.6,0,0}
\definecolor{deepgreen}{rgb}{0,0.5,0}
\definecolor{mybrickred}{RGB}{182,50,28}

\definecolor{fillcolor}{RGB}{216,217,252}

\usepackage{etoolbox}
\usepackage{framed}
\newif\ifxetexorluatex
\ifxetex
  \xetexorluatextrue
\else
  \ifluatex
    \xetexorluatextrue
  \else
    \xetexorluatexfalse
  \fi
\fi
\newcommand*\quotesize{60} 
\newcommand*{\openquote}
   {\tikz[remember picture,overlay,xshift=-4ex,yshift=-2.5ex]
   \node (OQ) {\fontsize{\quotesize}{\quotesize}\selectfont``};\kern0pt}

\newcommand*{\closequote}[1]
  {\tikz[remember picture,overlay,xshift=4ex,yshift={#1}]
   \node (CQ) {\fontsize{\quotesize}{\quotesize}\selectfont''};}

\colorlet{shadecolor}{white}

\newcommand*\shadedauthorformat{\emph} 

\newcommand*\authoralign[1]{%
  \if#1l
    \def\authorfill{}\def\quotefill{\hfill}
  \else
    \if#1r
      \def\authorfill{\hfill}\def\quotefill{}
    \else
      \if#1c
        \gdef\authorfill{\hfill}\def\quotefill{\hfill}
      \else\typeout{Invalid option}
      \fi
    \fi
  \fi}
{\authoralign{#1}
\ifblank{#2}
   {\def\shadequoteauthor{}\def\yshift{-2ex}\def\quotefill{\hfill}}
   {\def\shadequoteauthor{\par\authorfill\shadedauthorformat{#2}}\def\yshift{2ex}}
\begin{snugshade}\begin{quote}\openquote}
{\shadequoteauthor\quotefill\closequote{\yshift}\end{quote}\end{snugshade}}

\newfloat{Code}{htbp}{Code}

\usepackage{amsmath,amsfonts,bm}

\def\eqref#1{equation~(\ref{#1})}

\def\1{\bm{1}}

\DeclareMathAlphabet{\mathsfit}{\encodingdefault}{\sfdefault}{m}{sl}
\SetMathAlphabet{\mathsfit}{bold}{\encodingdefault}{\sfdefault}{bx}{n}

\newcommand\our{{MH-MoE}}

\newcommand\ourfull{{Multi-Head Mixture-of-Experts}}

\theoremstyle{plain}

\theoremstyle{definition}

\theoremstyle{remark}

\usepackage[textsize=tiny]{todonotes}

\definecolor{gray}{rgb}{0.5,0.5,0.5}
\definecolor{gray94}{gray}{.92}
\definecolor{gray90}{gray}{.90}
\definecolor{gray85}{gray}{.85}
\usepackage{xcolor}  
\usepackage{wrapfig}
\usepackage{colortbl}
\usepackage{pifont}
\usepackage{wrapfig}
\usepackage{epstopdf}
\usepackage{wrapfig}
\usepackage{cuted}
\usepackage{capt-of}

\title{{MH-MoE: Multi-Head Mixture-of-Experts}}

\author{
Shaohan Huang
~~~~Xun Wu~~~~Shuming Ma~~~~Furu Wei \\
Microsoft Research \\
{\href{https://aka.ms/GeneralAI}{https://aka.ms/GeneralAI}}
}

\begin{document}

\maketitle

\begin{abstract}
\ourfull{}~(\our{})~\cite{wu2024multiheadmixtureofexperts} demonstrates superior performance by using the multi-head mechanism to collectively attend to information from various representation spaces within different experts.
In this paper, we present a novel implementation of \our{}
that maintains both FLOPs and parameter parity with sparse Mixture of Experts models. 
Experimental results on language modeling tasks indicate that the new implementation yields quality improvements over both vanilla MoE and fine-grained MoE models. Additionally, our experiments show that \our{} is compatible with 1-bit Large Language Models (LLMs) such as BitNet~\cite{bitnet2}. 
\end{abstract}

\section{Sparse Mixture-of-Experts}
\label{sec:intro}

\label{sec:background}
Sparse Mixture-of-Experts (SMoE) provides a highly efficient way to scale neural network training and achieves better performance than dense models in various tasks~\cite{smoe,lepikhin2020gshard,du2021glam,kumatani2021building,chi2022representation, clark2022unified,zhao2023sparse, pham2023task}.
SMoE dynamically selects which parameters to use for each input, rather than applying the same parameters uniformly. This approach allows the networks to significantly increase the number of parameters while maintaining a roughly constant number of FLOPs per token.
Recent advancements in large language models employing Mixture of Experts (MoE) Transformers have demonstrated successful scaling to substantial sizes, accompanied by remarkable performance~\cite{Mixtral,dai2024deepseekmoe}. For instance, the Mixtral 8×7B, an SMoE model consisting of 8 experts (with activated 12.9 billion parameters), has been shown to outperform models such as LLaMA-70B.

In MoE architectures, the traditional Feed-Forward Networks (FFNs) within a Transformer are replaced by MoE layers. These MoE layers consist of multiple experts, each functioning as a standard FFN. The model employs a gating mechanism to route tokens to one or two of these experts per layer, utilizing either a top-1 or top-2 gating method.
The MoE layer consists of two components: $\mathbf{E}$ experts, each presented as $\text{Expert}_{i}: \mathbb{R}^{d} \rightarrow \mathbb{R}^{d}$, and a gate function, $G : \mathbb{R}^{d} \rightarrow \mathbb{R}^{\mathbf{E}}$. Given an input $\mathbf{x} \in \mathbb{R}^{d} $, the conditional output $\mathbf{y} \in \mathbb{R}^{d} $ is the weighted sum of gate function $G(\mathbf{x})$ and experts outputs $\{\text{Expert}_{i}(\mathbf{x})\}^{\mathbf{E}}_{i=0}$. The output $\mathbf{y}$ is computed by these activated experts, where $\Phi = \text{Top}_{k}\left(\text{Expert}_i\right)$ denote the set of activated experts and $|\Phi| = k$. 
\begin{equation}
\mathbf{y} = \sum_{p \in \Phi} G \left( \mathbf{x} \right) \cdot \text{Expert}_{p}\left(\mathbf{x}\right).
\end{equation}

\section{Multi-Head Mixture-of-Experts}

\subsection{Review of \ourfull{}}

Wu et al., \cite{wu2024multiheadmixtureofexperts} introduced \ourfull{} (\our{}), a novel approach that enhances the multi-head mechanism by enabling it to collectively attend to information from various representation spaces within different experts. \our{} incorporates two key modifications compared to the standard Sparse Mixture-of-Experts: adding a "heads" dimension $\mathbf{h}$ to the token dimension and ingratiating two linear projection layers at both the beginning and the end of the MoE layer.

Given an input $\mathbf{x} \in \mathbb{R}^{d}$, $d$ is the length of token dimension. First, $\mathbf{x}$ is projected by a linear layer with parameter matrices $\mathbf{W}_{\text{head}} \in \mathbb{R}^{d \times d}$,
\begin{equation}
    \hat{\mathbf{x}} =  \mathbf{x} \mathbf{W}_{\text{head}} 
    \label{EQ: multi-head}
\end{equation}
where $\hat{\mathbf{x}} \in \mathbb{R}^{d}$. After that, the token $\hat{\mathbf{x}}$ is split into $h$ sub-tokens along the token dimensions, and these sub-tokens are arranged in parallel according to the original token sequence, forming a new feature space $[\tilde{\mathbf{x}}_1,\tilde{\mathbf{x}}_2, ..., \tilde{\mathbf{x}}_h]$, where $ \tilde{\mathbf{x}}_h \in \mathbb{R}^{\frac{d}{h}}$ and $h$ denotes the number of heads.

Following the SMoE framework, the transformed input \(\tilde{\mathbf{x}}\) is fed into a MoE layer. This layer consists of \(\mathbf{E}\) experts, denoted as \(\text{Expert}_{i}: \mathbb{R}^{\frac{d}{h}} \rightarrow \mathbb{R}^{\frac{d}{h}}\), and a gating function \(G: \mathbb{R}^{\frac{d}{h}} \rightarrow \mathbb{R}^{\mathbf{E}}\). The output $\tilde{\mathbf{y}} \in \mathbb{R}^{\frac{d}{h}} $ is computed as following:
\begin{equation}
 \tilde{\mathbf{y}} = \sum_{p \in \Phi} G \left( \tilde{\mathbf{x}} \right) \cdot \text{Expert}_{p}\left(\tilde{\mathbf{x}}\right).
\end{equation}
where $\Phi$ is the set of activated experts. After processing through the MoE layer, all obtained outputs \(\tilde{\mathbf{y}}\) are rearranged into the original order of sub-tokens and concatenated together to form \(\hat{\mathbf{y}} \in \mathbb{R}^{d}\). This concatenated output \(\hat{\mathbf{y}}\) is then projected using a merge layer with parameter matrices \(\mathbf{W}_{\text{merge}} \in \mathbb{R}^{d \times d}\). This step ensures the effective integration of multiple features, capturing detailed information from different expert representation spaces.
\begin{equation}
    \mathbf{y} = \hat{\mathbf{y}} \mathbf{W}_{\text{merge}}  
    \label{EQ: merge}
\end{equation}
where $\mathbf{y}$ is the final output of the \our{} layer.

\subsection{Complexity Analysis}

We use \(\mathbf{B}\) to represent the number of tokens in batches, \(\mathbf{d}\) as the token dimension, \(\mathbf{d_{moe}}\) as the intermediate dimension in \(\text{Expert}(\mathbf{x})\), and \(\mathbf{h}\) as the number of multi-heads in \our{}. Assuming we use ``position-wise feed-forward networks'' (FFN)~\cite{transformer} in \(\text{Expert}(\mathbf{x})\), and opting for a version with no bias, \(\text{Expert}(\mathbf{x})\) can be computed as follows, where \( \mathbf{X} \in  \mathbb{R}^{\mathbf{B} \times \mathbf{\frac{d}{h}}}\), \(\mathbf{W_1} \in \mathbb{R}^{\mathbf{\frac{d}{h}} \times \mathbf{d_{moe}}}\) and \(\mathbf{W_2} \in \mathbb{R}^{ \mathbf{d_{moe}} \times \mathbf{\frac{d}{h}}}\):
\begin{equation}
    \text{Expert}(\mathbf{x}) = \text{FFN}_{\text{ReLU}}(\mathbf{X}, \mathbf{W_1}, \mathbf{W_2}) = \max(\mathbf{X} \mathbf{W_1}^{\top}, 0)\mathbf{W_2}
    \label{EQ: ffn}
\end{equation}

The number of scalar multiplications in \our{} is:
\begin{equation}
\overbrace{
2\mathbf{B}\mathbf{d}^2 - \mathbf{B} \mathbf{d}
}^{\text{Head Layer}}
+
\overbrace{
(4\mathbf{B}\mathbf{d}\mathbf{d_{moe}} - \mathbf{B} \mathbf{d} - \mathbf{B} \mathbf{d_{moe}} \mathbf{h}) \cdot k
}^{\text{Activated Experts}}
+
\overbrace{
2\mathbf{B}\mathbf{d}^2 - \mathbf{B} \mathbf{d}
}^{\text{Merge Layer}}
\label{EQ. sub-tokens}
\end{equation}

Assuming we use top-1 gating (set \(k = 1\)) and the intermediate dimension \(\mathbf{d_{moe}} = 4\mathbf{d}\), for sparse MoE, which does not include the head layer and merge layer, the number of scalar multiplications is \(16\mathbf{B}\mathbf{d}^2 - 5 \mathbf{B} \mathbf{d}\) and the leading term is \(16\mathbf{B}\mathbf{d}^2\).

In MH-MoE~\cite{wu2024multiheadmixtureofexperts}, they set the intermediate dimension \(\mathbf{d_{moe}} = 4\beta \mathbf{h} \mathbf{d}\), where \(\beta\) is a hyperparameter employed to scale the inner hidden dimension of FFNs. When the number of heads \(\mathbf{h}=4\) and \(\beta\) is \(\frac{63}{64}\) in their experiment, the scalar multiplications in ~\cite{wu2024multiheadmixtureofexperts} is \(67\mathbf{B}\mathbf{d}^2 - \frac{75}{4}\mathbf{B} \mathbf{d}\) and the leading term is \(67\mathbf{B}\mathbf{d}^2\). Although the activated parameters and whole model parameters in \cite{wu2024multiheadmixtureofexperts} are on par with sparse MoE, the FLOPS of \cite{wu2024multiheadmixtureofexperts} is significantly higher than the baseline. 

In our work, we will adjust the parameters in MH-MoE to maintain FLOPs parity with the vanilla method. Assuming the number of heads \(\mathbf{h} = 2\), we aim to keep the leading term at \(16\mathbf{B}\mathbf{d}^2\). To achieve this, we set the intermediate dimension \(\mathbf{d_{moe}} = 3 \mathbf{d}\) and increase the number of experts to match the model parameter count. Under this configuration, the number of scalar multiplications is \(16\mathbf{B}\mathbf{d}^2 - 6 \mathbf{B} \mathbf{d}\), ensuring that the leading term is on par with sparse MoE.

Alternatively, we can decrease the intermediate dimension to \(\mathbf{d_{moe}} = \frac{3}{2} \mathbf{d}\) and switch from top-1 gating to top-2 gating. This adjustment allows us to match not only the model parameters but also achieve parity in the number of scalar multiplications.

\subsection{Utilization Guidelines}

In this section, we will explain how to set the intermediate dimension \(\mathbf{d_{moe}}\) and the number of experts \(k\) in the mixture-of-experts layer. This process transforms a standard SMoE model into a MH-MoE model, ensuring that both the model parameters and the FLOPS are comparable to those of the standard SMoE model. The number of scalar multiplications in the sparse MoE is given by the following equation:
\begin{equation}
    (4\mathbf{B}\mathbf{d}\mathbf{d_{moe}} - \mathbf{B} \mathbf{d_{moe}} - \mathbf{B} \mathbf{d}) \cdot k
    \label{EQ. sparse-moe-FLOPS}
\end{equation}

Our goal is to ensure that the FLOPS of the MH-MoE model are equal to those of the standard SMoE model. We only consider the leading term of the equation, which is \(4\mathbf{B}\mathbf{d}\mathbf{d_{moe}} \cdot k\). 
From the equation~\ref{EQ. sub-tokens}, we can obtain the leading term of the FLOPS of the MH-MoE model is $4\mathbf{B}\mathbf{d}^2 + 4\mathbf{B}\mathbf{d}\mathbf{d_{mhmoe}} \cdot k$
The intermediate dimension \(\mathbf{d_{mhmoe}}\) can be set using the following equation:
\begin{equation}
    \mathbf{d_{mhmoe}} = \mathbf{d_{moe}} - \frac{\mathbf{d}}{k}
    \label{EQ. d-moe}
\end{equation}
where \(\mathbf{d_{moe}}\) is the intermediate dimension of the standard SMoE model, \(\mathbf{d}\) is the input dimension, and \(k\) is the number of experts. By setting the intermediate dimension \(\mathbf{d_{mhmoe}}\) using the equation~\ref{EQ. d-moe}, we can ensure that the FLOPs of the MH-MoE model are equal to those of the standard SMoE model.

As shown in equation~\ref{EQ. d-moe}, the MoE intermediate dimension of the MH-MoE model is smaller than that of the standard SMoE model. To maintain the same number of model parameters, we need to increase the number of experts \(\mathbf{E}\) in the mixture-of-experts layer.
The number of experts \(\mathbf{E}\) in the mixture-of-experts layer can be set using the following equation:
\begin{equation}
\begin{aligned}
    \overbrace{
    2\mathbf{d}\mathbf{d_{moe}} \cdot \mathbf{E_{moe}}
    }^{\text{\#parameter of standard MoE}} 
    & = 
    \overbrace{2\mathbf{d}^2 + 2\frac{\mathbf{d}}{\mathbf{h}}\mathbf{d_{mhmoe}} \cdot \mathbf{E_{mhmoe}}}^{\text{\#parameter of MH-MoE}} 
    \\ 
    & = 2\mathbf{d}^2 + 2\frac{\mathbf{d}}{\mathbf{h}}\left(\mathbf{d_{moe}} - \frac{\mathbf{d}}{k}\right) \cdot \mathbf{E_{mhmoe}}
    \label{EQ. expert-number}
    \end{aligned}
\end{equation}
This equation ensures that the number of parameters in the MH-MoE model matches that of the standard MoE model by appropriately adjusting the number of experts.

To illustrate with an example, let's assume \(\mathbf{d_{moe}} = 4\mathbf{d}\), we use top-1 gating (i.e., \(k = 1\)), and the number of heads is 3 (i.e., \(\mathbf{h} = 3\)). Using these values, we can derive the number of experts in the mixture-of-experts layer for the MH-MoE model.

First, recall the equation we derived for the intermediate dimension \(\mathbf{d_{mhmoe}} = \mathbf{d_{moe}} - \frac{\mathbf{d}}{k}\).
Substituting \(\mathbf{d_{moe}} = 4\mathbf{d}\) and \(k = 1\): $\mathbf{d_{mhmoe}} = 4\mathbf{d} - \frac{\mathbf{d}}{1} = 4\mathbf{d} - \mathbf{d} = 3\mathbf{d}$.
Next, we use the equation for parameter parity:
\begin{equation}
\begin{aligned}
    2\mathbf{d}\mathbf{d_{moe}} \cdot \mathbf{E_{moe}} 
    & = 2\mathbf{d}^2 + 2\frac{\mathbf{d}}{\mathbf{h}}\mathbf{d_{mhmoe}} \cdot \mathbf{E_{mhmoe}}
    \\ & = 2\mathbf{d}^2 + 2\frac{\mathbf{d}}{3}(3\mathbf{d}) \cdot \mathbf{E_{mhmoe}}
    \\ & = 2\mathbf{d}^2 + 2\mathbf{d}^2 \cdot \mathbf{E_{mhmoe}}
\end{aligned}
\end{equation}
Thus, the number of experts in the mixture-of-experts layer for the MH-MoE model is \(\mathbf{E_{mhmoe}} = 4\mathbf{E_{moe}} - 1\).

\section{Experiments}

\label{Sec: setup}
We adopt a decoder-only Transformer~\cite{gpt,gpt2} to evaluate the variants of MH-MoE and the baseline models on the RedPajama dataset~\cite{together2023redpajama}.  We use the same code base, training parameters, and pre-training tasks across all experiments. The decoder architecture comprises 12 layers with a model dimension of 768.

For the SMoE configuration, we employ top-1 gating with 8 experts, integrating MoE Transformer layers every two layers. The feedforward network utilizes SwiGLU~\cite{shazeer2020gluvariantsimprovetransformer}, with the intermediate dimension $\mathbf{d_{moe}}$ set to 2048.

We also implement a fine-grained version of the sparse MoE. In this configuration, the intermediate dimension is reduced to 1024, while the number of experts is increased to 16.

For MH-MoE, we compare two variants based on the number of heads, either 2 or 3. When the head number is 2, we set the intermediate dimension $\mathbf{d_{mhmoe}}$ to 768 and use top-2 gating to maintain FLOPs parity, increasing the number of experts to 40. For the variant with 3 heads, we set the intermediate dimension to 512, employ top-3 gating, and increase the number of experts to 96.

Furthermore, as employing a residual MoE setting, i.e., using shared experts~\cite{dai2024deepseekmoe}, has been shown to be effective in MoE models, we also conduct experiments under this setting to comprehensively validate the effectiveness of our MH-MoE. Specifically, a shared expert with the same size (hidden dimension is set to 2048) is applied to all MoE models.

\subsection{Language Modeling Evaluation}
For all experiments, we pre-train for 100,000 steps, with each training batch consisting of 0.5 million tokens. To evaluate the performance of different model architectures, we compute the perplexity on the validation set. Perplexity is reported at both 50,000 and 100,000 steps.

Table~\ref{tab:1} reports the results for MoE models without a shared expert, while Table~\ref{tab: residual} summarizes the results for MoE models incorporating a shared expert. Notably, across both settings, our MH-MoE consistently achieve lower perplexities compared to both the standard sparse MoE and its fine-grained variant. Additionally, the configuration with three heads outperforms the two-head configuration, demonstrating superior performance.

\begin{table*}[ht]
\centering
\caption{Validation set perplexity for the language modeling task. All models are matched in terms of parameters and computation.}
\begin{tabular}{lcccc}
\toprule
\textbf{Model}   & \multicolumn{1}{l}{\textbf{Training Steps}} & \textbf{RedPajama}      & \textbf{Wiki}           & \textbf{C4}             \\ \midrule
Dense            & \multirow{5}{*}{50,000}            & 13.01          & 12.95          & 17.41          \\
SMoE             &                                    & 11.87          & 10.51          & 15.63          \\
Fine-grained SMoE &                                    & 11.68          & 10.18          & 15.21          \\
MH-MoE (head=2)    &                                    & 11.60          & 10.11          & 15.11          \\
MH-MoE (head=3)    &                                    & \textbf{11.45} & \textbf{10.00} & \textbf{14.90} \\ \midrule
Dense            & \multirow{5}{*}{100,000}           & 12.13          & 11.58          & 16.21          \\
SMoE             &                                    & 10.90          & 9.68           & 14.35          \\
Fine-grained SMoE &                                    & 10.74          & 9.38           & 13.97          \\
MH-MoE (head=2)    &                                    & 10.70          & 9.26           & 13.80          \\
MH-MoE (head=3)    &                                    & \textbf{10.51} & \textbf{9.18}  & \textbf{13.63} \\ \bottomrule
\end{tabular}
\label{tab:1}
\end{table*}

\begin{table*}[ht]
\centering
\caption{Validation set perplexity for the language modeling task. All MoE models apply a shared expert~\cite{dai2024deepseekmoe} with the same size and matched in terms of parameters and computation.}
\begin{tabular}{lcccc}
\toprule
\textbf{Model}   & \multicolumn{1}{l}{\textbf{Training Steps}} & \textbf{RedPajama}      & \textbf{Wiki}           & \textbf{C4}             \\ \midrule
SMoE             &  \multirow{4}{*}{50,000}           & 11.76 & 10.33 &   15.19      \\
Fine-grained SMoE &                                     & 11.51 & 10.06 & 15.01        \\
MH-MoE (head=2)    &                                      & 11.48 & 9.91 & 14.87        \\
MH-MoE (head=3)    &                                      & \textbf{11.26} & \textbf{9.74} & \textbf{14.82}         \\ \midrule
SMoE             &  \multirow{4}{*}{100,000}        & 10.66 & 9.44 & 14.30        \\
Fine-grained SMoE &                                      & 10.41 & 9.15 & 13.78        \\
MH-MoE (head=2)    &                                      & 10.36 & 8.79 & 13.66        \\
MH-MoE (head=3)    &                                      & \textbf{10.28} & \textbf{8.72} & \textbf{13.49}        \\ \bottomrule
\end{tabular}
\label{tab: residual}
\end{table*}

\subsection{1-bit \our{}}
The recent impressive performance of BitNet~\citep{bitnet2} in quantizing and deploying large-scale models is heralding a new era for 1-bit Large Language Models (LLMs). Building on their impressive model performance, we conducted further experiments to explore whether our MH-MoE can effectively integrate with BitNet to achieve enhanced model optimization.

We employ the same experimental setting listed in Section~\ref{Sec: setup}, with the exception that all the models are quantized using BitNet. The corresponding experimental results are shown in Table~\ref{tab: bitnet}. In the 1-bit training and validation setting, we observed that our MH-MoE consistently outperformed other models, e.g., SMoE and Fine-grained SMoE. This demonstrates that MH-MoE integrates effectively with BitNet, enabling more lightweight deployment of MoE models without compromising performance. 

Besides, we observe a performance gap between the experimental results under the BitNet setting~(shown in Table~\ref{tab: bitnet}) and those under the non-BitNet setting~(shown in Table~\ref{tab:1}). We attribute this discrepancy to the fact that when the model size is relatively small, BitNet tends to degrade performance, a finding that aligns with the conclusions reported in the original BitNet paper~\cite{bitnet2}.

\begin{table*}[ht]
\centering
\caption{Validation set perplexity for the language modeling task. All dense and MoE models are quantized and trained using BitNet~\citep{bitnet2}, and matched in terms of parameters and computation.}

\begin{tabular}{lcccc}
\toprule
\textbf{Model} & \multicolumn{1}{l}{\textbf{Training Steps}} & \textbf{RedPajama} & \textbf{Wiki} & \textbf{C4}
\\ 
\midrule
Dense              & \multirow{5}{*}{50,000}              & 32.17 & 27.56 &  35.85    \\
SMoE               &                                      & 29.18  & 24.70 &  32.34 \\
Fine-grained SMoE  &                                      &  29.04 & 24.51  & 32.03       \\
MH-MoE (head=2)    &                                      & 28.84 & 24.27  &  31.86      \\
MH-MoE (head=3)    &                                      & \textbf{28.77} & \textbf{24.13}  &  \textbf{31.81}       \\ 
\midrule
Dense            & \multirow{5}{*}{100,000}            &        30.04   &       24.75    & 33.55         \\
SMoE             &               & 26.78 & 21.54 &  29.73       \\
Fine-grained SMoE &                                       & 26.68 & 21.42  &  29.50     \\
MH-MoE (head=2)    &                                      & 26.59 & 21.11  &  29.27     \\
MH-MoE (head=3)    &                                      & \textbf{26.47} & \textbf{21.06}  &  \textbf{29.14}      \\ \bottomrule
\end{tabular}
\label{tab: bitnet}
\end{table*}

\subsection{Ablations}
In this section, we conduct a detailed ablation study focusing on the head layer and the merge layer, both of which are integral components of~\our{}. The design of these layers draws inspiration from the multi-head attention mechanism ~\cite{transformer}.
Specifically, in our Multi-Head Mixture-of-Experts model, we conceptualize the head layer as constituting the query, key, and value projections. The merge layer, on the other hand, is considered the output projection.
It is crucial to thoroughly investigate their contributions and understand their impact

We separately integrate head and merge layers into both our baseline SMoE and fine-grained SMoE models. Table~\ref{tab:2} presents the validation set perplexity for various models with and without these layers. It is important to note that all models without the head and merge layers maintain the same number of scalar multiplications, and similarly, all models with the head and merge layers also maintain an equivalent number of scalar multiplications.

Our findings indicate that for both the SMoE and fine-grained SMoE models, the addition of head and merge layers—which inevitably increases the number of FLOPs in these layers—results in only marginal gains in performance. In contrast, for the \our{} model, the inclusion of the head and merge layers leads to significant improvements in performance. This underscores the critical role these layers play in enhancing the effectiveness of the \our{} model.

\begin{table*}[ht]
\centering
\caption{Validation set perplexity for different models with and without head and merge layers.}
\begin{tabular}{lcccc}
\toprule
\textbf{Model}    & \textbf{w/ head \& merge layer} & \textbf{RedPajama}      & \textbf{Wiki}           & \textbf{C4}             \\ \midrule
SMoE              & \ding{55} & 11.87 &	10.51 & 	15.63 \\
SMoE              & \ding{51} & 11.84 &	10.48 & 	15.61 \\ \midrule
Fine-grained SMoE & \ding{55} & 11.68 &	10.18 & 	15.21 \\
Fine-grained SMoE & \ding{51} & 11.67 &	10.18 & 	15.19 \\ \midrule
MH-MoE (head=2)   & \ding{55} & 11.71 &	10.16 & 	15.23 \\
MH-MoE (head=2)   & \ding{51} & 11.46 &	9.98	& 14.89 \\ \bottomrule
\end{tabular}
\label{tab:2}
\end{table*}

We further analyze the head and merge layers separately. As shown in Table~\ref{tab:3}, both of these layers contribute positively to model performance. Notably, the head layer provides a more substantial gain compared to the merge layer. This suggests that while both layers are beneficial, the head layer plays a more critical role in enhancing model effectiveness.

\begin{table*}[ht]
\centering
\caption{Validation set perplexity for ablation of head and merge layers.}
\begin{tabular}{ccccc}
\toprule
\textbf{w/ head layer}    & \textbf{w/ merge layer} & \textbf{RedPajama}      & \textbf{Wiki}           & \textbf{C4}             \\ \midrule
 \ding{55} & \ding{55} & 11.97 &	10.40 & 	15.52 \\
 \ding{51} & \ding{55} & 11.74 &	10.18 & 	15.17 \\
 \ding{55} & \ding{51} & 11.84 &	10.27 & 	15.36 \\
 \ding{51} & \ding{51} & 11.60 &	10.11 & 	15.11 \\  \bottomrule
\end{tabular}
\label{tab:3}
\end{table*}

Through our ablation experiments, we aim to dissect the individual contributions of the head and merge layers. By systematically altering or removing components within these layers, we can gain insights into how each part influences the overall model performance. This analysis not only helps in validating our design choices but also provides guidance for potential improvements and optimizations in future iterations of the model.

\section{Conclusion}

In this work, we present a new implementation of \our{} to ensure FLOPs parity with sparse Mixture of Experts (MoE) models. Our experimental results show that the new variants both outperform both vanilla SMoE models and fine-grained MoE models under various  experimenta settings. Additionally, we conducted ablation experiments to analyze the impact of head and merge layers. We demonstrate that both head and merge layers improve model performance, with the head layer yielding particularly substantial gains.

\bibliographystyle{alpha}
\bibliography{anthology}

\end{document}